\documentclass{article}

\usepackage{arxiv}

\usepackage[utf8]{inputenc} 
\usepackage[T1]{fontenc}    
\usepackage{hyperref}       
\usepackage{url}            
\usepackage{booktabs}       
\usepackage{amsfonts}       
\usepackage{nicefrac}       
\usepackage{microtype}      
\usepackage{lipsum}
\usepackage{graphicx}
\usepackage{subcaption}
\usepackage{tabularx}
\usepackage{graphicx}
\usepackage{latexsym}
\usepackage{times}
\usepackage{soul}
\usepackage{url}
\usepackage{amsmath}
\usepackage{amssymb}
\usepackage{amsthm}
\usepackage{booktabs}
\usepackage{algorithm}
\usepackage{algorithmic}
\usepackage{natbib}

\def\x{\mathbf{x}}

\title{Automatic Algorithm Selection for Pseudo-Boolean Optimization with Given Computational Time Limits}

\author{
 Catalina Pezo \\
 Department of Computer Science\\
 Universidad de Concepci\'on\\
 Concepci\'on, Chile \\
\texttt{cpezo2017@inf.udec.cl} \\
   \And
 Dorit Hochbaum \\
 Department of Industrial Engineering and Operations Research\\
 University of California, Berkeley\\
 Berkeley, CA, USA \\
\texttt{dhochbaum@berkeley.edu} \\
\And
 Julio Godoy \\
 Department of Computer Science\\
 Universidad de Concepci\'on\\
 Concepci\'on, Chile\\
\texttt{jugody@udec.cl} \\
  \And
  Roberto As\'in-Ach\'a\\
  Department of Computer Science\\
  Universidad T\'ecnica Federico Santa Mar\'ia\\
  San Joaqu\'in, RM, Chile
 \texttt{roberto.asin@usm.cl} \\
}

\begin{document}
\maketitle
\begin{abstract}
Machine learning (ML) techniques have been proposed to automatically select the best solver from a portfolio of solvers, based on predicted performance. These techniques have been applied to various problems, such as Boolean Satisfiability, Traveling Salesperson, Graph Coloring, and others.

These methods, known as meta-solvers, take an instance of a problem and a portfolio of solvers as input. They then predict the best-performing solver and execute it to deliver a solution. Typically, the quality of the solution improves with a longer computational time. This has led to the development of {\em anytime} selectors, which consider both the instance and a user-prescribed computational time limit. {\em Anytime meta-solvers} predict the best-performing solver within the specified time limit.

Constructing an anytime meta-solver is considerably more challenging than building a meta-solver without the ``anytime" feature. In this study, we focus on the task of designing anytime meta-solvers for the NP-hard optimization problem of {\em Pseudo-Boolean Optimization} (PBO), which generalizes Satisfiability and Maximum Satisfiability problems. The effectiveness of our approach is demonstrated via extensive empirical study in which our anytime meta-solver improves dramatically on the performance of Mixed Integer Programming solver Gurobi, which is the best-performing single solver in the portfolio. For example, out of all instances and time limits for which Gurobi failed to find feasible solutions, our meta-solver identified feasible solutions for $47\%$ of these.
\end{abstract}


\section{Introduction}

\textit{Per-instance Automatic Algorithm Selection} (AAS), first proposed in \citep{rice1976algorithm}, consists of, for a given instance of a known problem and a portfolio of algorithms for the problem, a prediction of an algorithm in the portfolio that best solves the given instance. The prediction is done by Machine Learning models that are trained on a set of problem instances. This is of particular interest for NP-hard optimization problems since, for such problems, there is no single algorithm that dominates the others on every instance in every possible scenario. The \textit{anytime behavior} of an algorithm, when a feasible solution is available, is its profile of improvement in the objective function value at each successive time step. \textit{Anytime Automatic Algorithm Selection} aims to choose the algorithm which is expected to find the best possible solution, within the given time limit, for a specific instance. Previously, Anytime Automatic Algorithm Selection meta-solvers were proposed for the Knapsack \citep{huerta2020anytime} and Traveling Salesperson \citep{huerta2022improving} problems.


We devise here Anytime Automatic Algorithm Selection for the NP-hard Pseudo-Boolean Optimization problem (PBO) \citep{boros2002pseudo}.  PBO is an optimization problem with an objective function that is a Pseudo-Boolean function and subject to constraints that are (in)equalities over Boolean variables. 
Many problems are typically modeled as PBO, including hardware and software verification \citep{manquinho2005effective, wille2011atpg}, software dependency \citep{trezentos2010apt}, planning \citep{acha2022multi}, scheduling problems \citep{asin2014curriculum}, and the satisfiability problem (SAT), the maximum satisfiability problem (MaxSAT) \citep{biere2009handbook}.  As such, improving the ability to deliver high-quality solutions for PBO can impact the solvability of a broad range of problems.
Indeed, a number of commercial and publicly available algorithms (solvers) have been proposed for PBO and the SAT community maintains the Pseudo-Boolean Competition \citep{manquinho2011pseudo} in which the performance of state-of-the-art solvers is assessed.


This paper describes a meta-solver that, for a given instance and time limit, i) predicts, using a Machine Learning model, which solver, among a portfolio of solvers, will deliver a best quality (smallest objective value) feasible solution and ii) executes such a solver. 
Our experiments demonstrate that our meta-solver outperforms all the individual solvers in the portfolio by a wide margin. In particular, our meta-solver outperforms Gurobi -- which is the dominant solver in the portfolio -- in achieving better quality solutions, for a portion of the instances and time limits where Gurobi finds feasible solutions. And in 47\% out of the cases where Gurobi does not identify feasible solutions, our meta-solver does find feasible solutions.
A major contribution of our meta-solver is that identifies with great precision when feasibility is \textit{not} expected to be attained for the given instance, within the specified time limit. 

Beyond achieving improved results, our study provides insights into the most important features that determine the choice of the best solver. We identify the fraction of the number of terms that appear on the objective function, out of the total number of terms in the objective and the constraints, as a major feature. This feature has not appeared previously in algorithm selection studies on SAT and MaxSAT. Another major feature is the prescribed time limit, which appears to be more important than other characteristics of the instances in determining the solver selection. 

The paper is organized as follows:  In Section~\ref{sec:related} we discuss related work. Section~\ref{sec:prelims} presents essential concepts and terminology.  Section~\ref{sec:aaas-pbo} describes the meta-solver construction and Section~\ref{sec:results} presents and analyzes experimental results. Finally, Section~\ref{sec:conclusions} discusses future work and conclusions.

\section{Related Work}
\label{sec:related}

This section provides an overview of related work on Pseudo-Boolean Optimization solvers, and on Automatic Algorithm Selection for the 
SAT and MaxSAT problems, which are special cases of PBO. 

To facilitate the reading of the paper we provide, in Table~\ref{tab:acronyms}, a list of acronyms used throughout. 

\begin{table}[hhh!]
\footnotesize
    \centering
    \begin{tabular}{cc|cc}
    \textbf{Acronym} & \textbf{Definition} & \textbf{Acronym} & \textbf{Definition}\\
    \hline
    AAS   & Automatic Algorithm Selection & LSU   & Linear SAT-UNSAT algorithm\\
    AAAS  & Anytime Automatic Algorithm Selection &  MaxSAT & Maximum Boolean Satisfiability problem\\
    ASLib & Algorithm Selection Library & ML    & Machine Learning\\
    ASP   & Answer Set Programming & MIP   & Mixed Integer Programming\\
    BCS  & Boolean Constraint Satisfaction & NaPS  & Nagoya Pseudo-Boolean Solver\\
    BDD   & Binary Decision Diagram & PB    & Pseudo-Boolean\\
    CDCL & Conflict Driven Clause Learning & PBO   & Pseudo-Boolean Optimization\\
    GB   & Gradient Boosting & RF & Random Forest \\
    CNF   & Conjunctive Normal Form & SAT & Boolean Satisfiability Problem\\
    CNN   & Convolutional Neural Network &SBS   & Single Best Solver\\
    KNN   & K-Nearest Neighbors &  VBS   & Virtual Best Solver\\
    LP    & Linear Programming & WBO  & Weighted Boolean Maximization\\
    LS    & Local Search & WPM  & Weighted Partial MaxSAT\\
   \hline    
    \end{tabular}
    \caption{Acronyms used in this paper}
    \label{tab:acronyms}
\end{table}

\subsection{PBO solvers}
\label{sec:solvers}
Most PBO solvers are based on making calls to a program subroutine, based on the Conflict-Driven Clause-Learning (CDCL) algorithm \citep{biere2009conflict}, that solves a decision problem on whether the input formula is feasible or not. The optimization problem is translated into a feasibility problem by adding to the constraints the \textit{objective function constraint}, which is an inequality specifying that the objective function is less than or equal (for minimization) to a specified upper bound. This translates the PBO problem into a Boolean Constraint Satisfaction (BCS) problem. Many solvers (e.g. \citep{sorensson2010minisat, sakai2015construction,martins2014incremental}) further encode the BCS problem as a CNF Satisfiable (SAT) formula. Another family of solvers, e.g. \citep{wolsey1999integer,gurobi}, implement a Branch \& Bound search strategy on a search tree that, at each node, solves the linear relaxation of the problem. In addition to these, a third family of solvers uses local search procedures \citep{lei2021efficient}. 

Next, we list the PBO solvers considered for inclusion in the portfolio of the meta-solver. These solvers were chosen due to their good performance in the PBO competitions \citep{manquinho2011pseudo}.

\begin{description}
    \item[NaPS:] The Nagoya Pseudo-Boolean Solver \citep{sakai2015construction} won the 2016 Pseudo-Boolean Competition in $4$ categories. This solver is a MaxSAT solver, based in Minisat+\citep{sorensson2010minisat}. The main difference between NaPS and other PBO solvers that translate the formula to MaxSAT, is that NaPS uses Binary Decision Diagrams (BDD) to translate the PB constraint to a SAT formula.
    \item[OpenWBO:] Open-WBO \citep{martins2014open} is a weighted partial MaxSAT solver that won second place in two categories in the 2016 PBO Competition. PBO instances are easily translated into weighted partial MaxSAT instances where the PBO's constraints are translated into hard clauses (that must be satisfied), and the objective function is translated into a set of weighted soft clauses. Open-WBO implements five different search algorithms, of which we only consider two since the other three were dominated by other algorithms in our portfolio. The two search algorithms are:
    \begin{description}
    \item[Linear-su:] This algorithm translates the PBO instance to Weighted Partial MaxSAT and uses the LSU search strategy as explained in \citep{koshimura2012qmaxsat}. We will refer to this option as  {\em OpenWBO-lsu}. 
    \item[oll:] 
    This algorithm translates the PBO instance into Weighted Partial MaxSAT and uses a search strategy similar to WPM1, as explained in \citep{ansotegui2012improving}. This option will be referred to as {\em OpenWBO-oll}.
    \end{description}
    \item[Clasp:] Clasp \citep{gebser2007clasp} is part of the PosTdam Answer Set Solving COllection, POTASSCO. It is a CDCL solver for Answer Set Programming. Answer Set Programming (ASP) is a form of declarative programming oriented towards difficult (primarily NP-hard) search problems that is more expressive and subsumes PBO. It uses different semantics than other CDCL solvers and, as such, it has superior performance for certain subsets of instances.
    \item[LS-PBO:] The local search LS-PBO solver achieved good performance in instances from the PB competition. It features a transformation of the objective function into objective constraints, a constraint weighting scheme for the Pseudo-Boolean constraints, and a scoring function to guide the local search \citep{lei2021efficient}.
    \item[Gurobi:] Gurobi \citep{gurobi} is a Mixed Integer Programming (MIP) commercial solver that is able to handle mixed linear, quadratic and second-order cone constraints. When solving a PBO instance, Gurobi uses a Branch \& Bound search procedure powered by advanced preprocessing techniques, intelligent generation of cutting planes, specialized heuristics, and parallel processing. Here we used version 9.5.0.  
    \item[RoundingSat:] The RoundingSat solver, originally introduced in \citep{elffers2018divide}, is a CDCL solver that includes faster propagation routines for PB constraints. Unlike other solvers, it does not translate the PB constraints into a SAT formula but executes conflict analysis directly on the PB constraints.
    It also allows for incorporating a Linear Programming (LP) solver into its pipeline.
    
\end{description}

\subsection{Algorithm Selection for SAT and MaxSAT}
\label{sec:aasat}
A thorough review of Automatic Algorithm Selection (AAS) is provided in \citep{kerschke2019automated}. 
The performance of AAS meta-solvers has been improved over time due to the influence of algorithm selection competitions \citep{lindauer2019algorithm} and the maintenance and updating of the Algorithm Selection Library (ASLib) \citep{bischl_aslib_2016}. 

In particular, for SAT and MaxSAT (which are closely related to PBO), many successful meta-solving approaches were proposed in \citep{xu2008satzilla}, \citep{malitsky2012parallel}, \citep{ansotegui2016maxsat},\citep{hoos2015aspeed}, \citep{pulina2007multi}, \citep{gebser2011portfolio}, \citep{maratea2014multi}. For example, the SATzilla solver \citep{xu2008satzilla} has been quite influential in the SAT community and won several categories in different versions of the SAT competition and SAT evaluation. SATzilla is a Portfolio-Based Algorithm Selection system that chooses the appropriate solver in the portfolio, based on the computation of a number of features from the input instance and other features it collects from probing procedures. For MaxSAT, an improved instance-specific algorithm configuration, also based on different formula and probing features, was proposed in \citep{ansotegui2016maxsat}. This solver won the majority of the categories of the MaxSAT competition in 2016.

SATzilla's first version \citep{nudelman2004satzilla} proposed $84$ features for characterizing SAT instances, classified into $9$ categories: problem size, variable-clause graph, variable graph, clause graph, balance, proximity to Horn formulae, LP-based, CDCL probing and local search probing features. \citep{xu2008satzilla} used $48$ of those proposed features, excluding the computationally expensive ones, to build SATzilla. In \citep{ansotegui2016maxsat}, $32$ of the standard SAT features were selected, such as the number of variables, number of clauses, proportion of positive to negative literals, and average number of clauses in which a variable appears, among others. For the specific MaxSAT problem, they also computed the percentage of clauses that are soft and the statistics of the distribution of weights. 

In \citep{loreggia2016deep}, the authors propose a new approach to AAS, following the philosophy of deep learning models that replace domain-specific features with generic raw data, from which they learn the important features automatically. For this, the authors propose to use as raw data the input text file of any combinatorial problem and convert it to a fixed-size image, that will be used as input for a Convolutional Neural Network (CNN). Specifically, they first create a vector from the input file, replacing each character with its ASCII code, they then reshape the vector as a matrix of $\sqrt{\textit{N}} \times \sqrt{\textit{N}}$, where \textit{N} is the number of total characters in the input text file. Finally, this new ``image" of ASCII values is re-scaled to a predefined size, to work with a set of images of the same size. With this input, the selector is a trained CNN multi-label classification model that encodes the input instance and outputs the most promising solver for the instance. This approach is tested with SAT and Constraint Satisfaction (CSP) instances, obtaining a meta-solver that is able to outperform the Single Best Solver (see Subsection~\ref{sub:aas}), but underperforms in comparison with methods based on domain-specific features. As a baseline for our work, we will use a straightforward adaptation to this approach to anytime scenarios, since no specific work on Anytime Automatic Algorithm Selection for PBO has been proposed until now.

\section{Preliminaries}
\label{sec:prelims}

In this Section, we give an overview of Machine Learning methods that we use and provide formal definitions of the Pseudo-Boolean Optimization problem, as well as the Automatic Algorithm Selection problem. We also present a performance metric, called the $\hat{m}$, that is used in addition to accuracy and confusion matrix to assess the performance of the proposed meta-solver.

\subsection{Machine Learning}
\label{sub:prelims-ml}
Over the past decade, the field of Machine Learning (ML) within Artificial Intelligence has undergone significant development, according to \citep{alpaydin2021machine}. ML has become a powerful tool for processing and analyzing large volumes of data, as algorithms developed for various ML models aim to uncover hidden patterns within the data.  These models learn from a given set of data, called a training set, to create a function $f$ that maps an input instance to a corresponding scalar or vector output, referred to as labels.

The process by which $f$ is learned determines the classification of the ML model. If the learning process relies on ground truth labels, consisting of input instances and their corresponding output labels, then the model is considered supervised. Examples of supervised models can be found in Burkart's survey~\citep{burkart2021survey}. If the model finds patterns independently without access to ground truth labels, it is considered unsupervised, as seen in Alloghani's work~\citep{alloghani2020systematic}. Semi-supervised models combine ground truth labels with pattern analysis of input data to learn, as described in \citep{zhu2009introduction}.

Supervised and unsupervised machine learning can both perform automatic algorithm selection/configuration, as demonstrated by the work of~\citep{kadioglu2010isac}. However, the focus of this text is on supervised Machine Learning.

ML models can also be classified based on the nature of the output produced by $f$. If the output consists of discrete values used to categorize inputs into different classes, the model is considered a classification model. On the other hand, if the output corresponds to real values, the model is a regression model.

The supervised ML algorithms for classification used here are:
\begin{description}
\item[Random Forest:] The Random Forest (RF) method, as described by Breiman~\citep{breiman2001random}, is an ensemble technique~\citep{breiman1996bagging} that constructs a specified number of decision trees (controlled by a parameter $n_{estimators}$). Each tree is trained on a different subset of instances within the training set and proposes a result to compute the output label. The final output label is determined through a consensus scheme that differs depending on whether the model is a regression or classification model.

In the case of regression, the consensus is reached by averaging the outputs of all the decision trees. In contrast, for classification models, the output label corresponds to the most frequently repeated label (voted) among the decision trees' outputs. The decision trees themselves are constructed using the decision tree algorithm described in~\citep{quinlan1986induction}.

\item[$k$-Nearest Neighbors:] The $k$-Nearest Neighbors (KNN) algorithm, introduced by Fix and Hodges~\citep{fix1991discriminatory}, is a Machine Learning (ML) method that determines the output label based on the labels of the $k$ closest training examples to the input point being labeled. The distance between the feature vectors of the input point and the training examples can be calculated using various metrics, but the most commonly used is the Euclidean distance.

For classification tasks, KNN assigns the output label as the most frequently occurring label among the $k$ neighbors. In the case of regression tasks, the output label corresponds to the average of the labels of the $k$ neighbors.

\item[Gradient Boosting:] Gradient Boosting (GB) is an ML method, proposed in~\citep{friedman2001greedy}, that builds upon the ideas behind Ada Boost. GB allows for different parameterized loss functions to be defined. The learning process involves consecutively training a parameterized number ($n_{estimators}$) of new ``weak" models, with each new model being given as input to the next iteration.

In a manner similar to gradient descent, a negative gradient is computed based on the past model, which is weighted according to a parameterized scheme ($learning\_rate$). A move in the opposite direction is then taken to reduce the loss. This process is repeated to improve the performance of the model.
\end{description}

\subsection{Pseudo-Boolean Optimization (PBO)}
\label{sec:pbo}

A Pseudo-Boolean function is a mapping $f :\{0,1\}^{n} \rightarrow \mathbb{R}$, where $\mathbb{R}$ is the set of real numbers, 
\citep{boros2002pseudo}. A Pseudo-Boolean Optimization Problem (PBO) is formulated for an array of Boolean variables $\x$ as follows:
\begin{eqnarray*}
\min\ & &f(\x)\\
\text{s.t. }
   & & g_1(\x)  \geq a_1\\
   & & \vdots  \\
   & & g_n(\x)  \geq a_n\\
   & & \x\in \{0,1\}^n.
\end{eqnarray*}

\noindent  Without loss of generality, the constraints are of the form $g_i(\x) = b_1 t_1 + b_2 t_2 + \ldots + b_n t_m $, where $b_i$ are integers, and $t_j$, called a {\em term}, is a product of the variables in a subset $S_j \in \{1,2,\ldots,n\}$, $t_j = \prod_ {k\in S_j} x_k$.

\subsection{Performance metric for Anytime Automatic Algorithm Selection}
\label{sub:aas}
A specific type of Algorithm Selection is {\em per-instance Automatic Algorithm Selection (AAS)}: given a problem $P$, with $I$ a set of instances of $P$, $A=\{A_1,A_2,...,A_n\}$ a set of algorithms for $P$ and a general given metric $pm$ that measures the performance of any algorithm $A_j \in A$ for $I$, AAS consists of a selector $S$ that maps any instance $i \in I$ to an algorithm $S(i) \in A$ such that the overall performance of $S$ on $I$ is optimal according to metric $pm$. 

In order to measure the performance of a solver $s \in A$ over time, we discretize the time-space into \textit{timesteps}.
Let $I$ be a set of instances, and $T$ a set of timesteps.  For the instance-timestep pair $(i,t) \in I \times T$, let $o_s(i,t)$ be the objective value of $s$ on instance $i$, at timestep $t$. Since the value for $o_s(i,t)$ can greatly vary across instances and timesteps, in order for each data point to weigh equally in a cumulative metric, a normalization function $n(o_s(i,t),i,t)$ is used to map $o_s(i,t)$ values to a uniform range. For PBO, we use the normalization given in~(\ref{eq:n}). The cumulative metric $m_s$ we use, also considered in \citep{amadini2014sequential}, is defined as:
\begin{equation}
\label{eq:m-s}
    m_s = \sum_{(i,t) \in I \times T} n(o_s(i,t),i,t)
\end{equation}

\noindent and corresponds to the normalized cumulative performance of solver $s$ across all pairs $(i,t) \in I \times T$. For a meta-solver $ms$ that for each (i,t) instance-timestep pair selects solver $s'_{i,t}$ its cumulative performance metric is defined as:
\begin{equation}
\label{eq:m-ms}
    m_{ms} = \sum_{(i,t) \in I \times T} n(o_{s'_{i,t}}(i,t),i,t)
\end{equation}

The evaluation of meta-solvers is usually done in comparison to the performance of single best solver and virtual best solver, defined as:
\begin{description}
    \item[Single Best Solver (SBS):] The single algorithm that performs best (on average) on {\em all} instances.
    \item[Virtual Best Solver (VBS):] A solver that makes perfect decisions and matches the best-performing algorithm for each problem instance, without overhead.
\end{description}

For an algorithm selector meta-solver $ms$, the $\hat{m}_{ms}$ metric was proposed for the Algorithm Selection Competitions \citep{lindauer2019algorithm}, using, for each solver $s$, the performance metric $m_s = \sum_{i \in I } n(o_s(i),i)$. Here we generalize it for anytime algorithm selector meta-solvers as follows. 
\begin{equation}
\label{eq:m-hat}
    \hat{m}_{ms} = \dfrac{m_{ms} - m_{VBS}}{m_{SBS}- m_{VBS}}
\end{equation}

\noindent where $m_{ms}$ is the normalized cumulative performance of meta-solver $ms$, $m_{VBS}$ is the normalized cumulative performance of the VBS, and $m_{SBS}$ is the normalized cumulative performance of the SBS.

We observe that:
\begin{itemize}
    \item The closer $\hat{m}_{ms}$ to $0$, the more similar the meta-solver is to the VBS.
    \item If $\hat{m}_{ms} > 1$, then the meta-solver is worse than the SBS and, hence, is not useful.
\end{itemize}

\section{Designing the Machine Learning Oracle for AAAS for PBO}
\label{sec:aaas-pbo}

In this section we describe the workflow we carried on for designing and implementing AAAS Machine Learning oracles for PBO. In Subsection~\ref{sec:pbo-inst-solvers} we give details on how we recorded the anytime behavior of the solvers on the chosen portfolio and elaborate on the characteristics of the instance benchmarks used for our work. Subsection~\ref{sec:pbo-anytime} presents the dataset we used for the training and testing of our meta-solver, and Subsection~\ref{sub:characterization} describes the possibilities we considered for characterizing the instances to use as input features to our models. Further details on the ML models and algorithms tested can be found in Subsection~\ref{sec:eval}. Finally, Subsection~\ref{sec:eval} presents the evaluation of the implemented ML models.


\subsection{PBO instances and solvers}
\label{sec:pbo-inst-solvers}

The dataset of PBO instances was obtained from the 2006, 2007, 2009, 2010, 2011, 2012, 2015, and 2016 Pseudo-Boolean Competitions \citep{manquinho2011pseudo}. These instances were collected from different domain applications such as Bio-informatics, Timetabling, and Hardware verification, among others. The instances with similar origins are organized in benchmarks or ``families", and differ from one another in the type of constraints (linear or nonlinear) and the magnitudes of the constraints' coefficients (normal integers or arbitrary precision integers). Our experimental study includes $118$ benchmarks, for a total of $3128$ \emph{feasible} instances.

The solvers used for the construction of the meta-solver, described in detail in Section \ref{sec:solvers}, are: NaPS, two variants of OpenWBO, LS-PBO, RoundingSat, Gurobi and Clasp. We only considered solvers that either have their codes available so as to modify them to record their anytime behavior, or already provide this capability by default.

\begin{figure}[!htb]
    \centering
    \includegraphics[width=1\columnwidth]{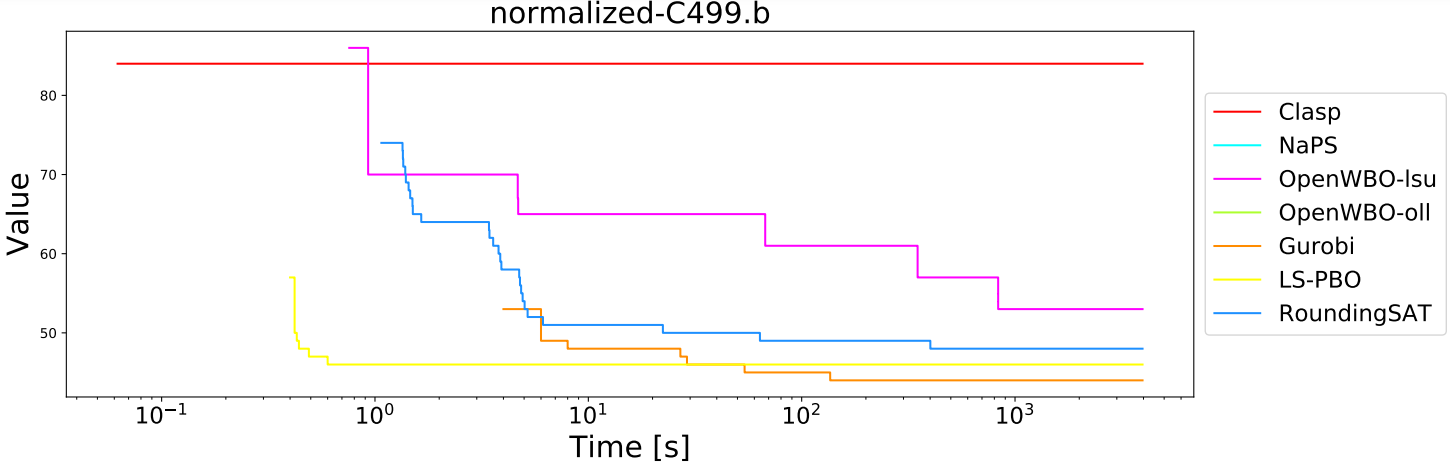}
    \caption{Anytime behavior of the solvers for the instance  ``normalized-C499\_b" from Benchmark 106.} 
    \label{results}
\end{figure}

In order to evaluate the anytime behavior of the solvers, we discretize a time interval of one hour into $500$ timesteps following a logarithmic scale, analogous to \citep{huerta2022improving}. We keep track, whenever the value of the objective function improves, of the corresponding timestep, and the updated new solution.

The anytime behavior of each solver is recorded as the updated best objective value (incumbent) for each of the $500$ timesteps. Figure~\ref{results} shows the anytime behaviors of the solvers for the instance ``normalized-C499\_b", which corresponds to a logic-synthesis application. Note that the solver that outputs the solution with the smallest value at a given timestep $ts$ is considered the best option for any specified time limit between the time corresponding to $ts$ and the next timestep $ts+1$. In Figure~\ref{results} we observe the change in the best solver across the timeline. Initially, for small time limits, \textit{Clasp} is the best solver. Then, after a few milliseconds, \textit{LS-PBO} becomes the best solver, but it is finally outperformed by \textit{Gurobi}.

A solver is said to \textit{win} an instance-timestep pair if it computes the best-found solution (i.e. a feasible solution with the best objective value) for that instance in that timestep. Ties are broken in favor of the solver that achieved such best incumbent first.

\subsection{Training and testing dataset generation}
\label{sec:pbo-anytime}

\begin{figure}[h]
    \centering
    \includegraphics[width=1\columnwidth]{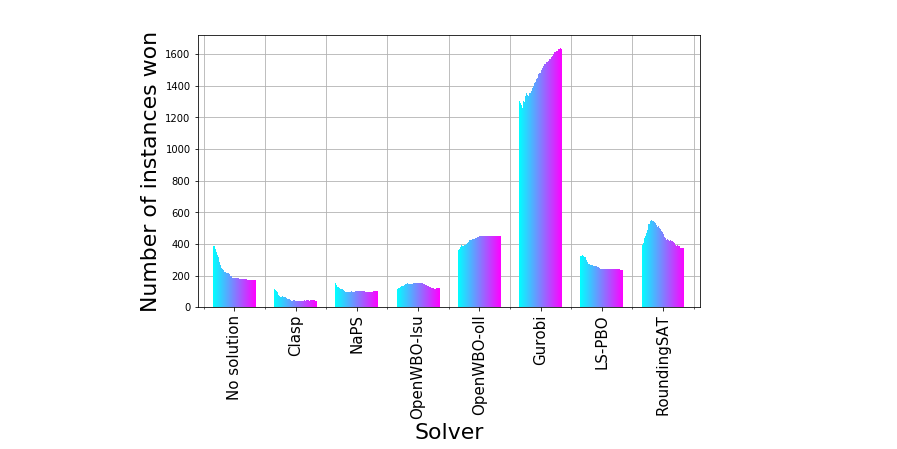}
    \caption{Number of wins for each solver across the time horizon (see explanation in text).}
    \label{bars}
\end{figure}

\begin{figure}[h]
    \centering
    \includegraphics[width=0.9\columnwidth]{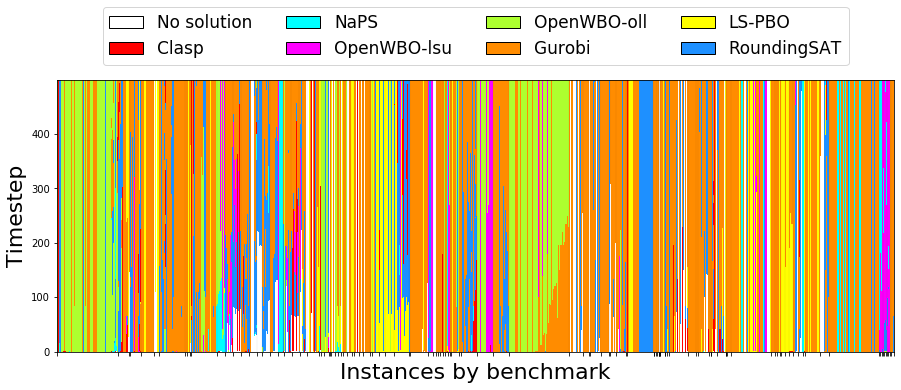}
    \caption{Best solver per instance for each timestep. The horizontal axis represents  the $3128$ instances arranged in the $118$ benchmarks. Each vertical bar displays, for one instance, the change in the best solver over the timesteps.}
    \label{matrix}
\end{figure}

The dataset we used for building our ML oracles was generated from running all the solvers on the portfolio over all the instances we collected. Figure~\ref{bars} summarizes the number of wins for each solver across the time horizon, for each of the $3128$ instances. For each solver, on the horizontal axis, there is a bar consisting of $500$ vertical lines, colored from light blue (for the small timesteps) all the way to purple (for the large timesteps). We include a ``no solution'' entry for instance-timestep pairs where no feasible solution was identified by any of the solvers. Throughout various instances and time intervals, four dominant solvers emerge: Gurobi, RoundingSAT, OpenWBO-oll and LS-PBO. RoundingSAT and LS-PBO exhibit a greater share of wins for smaller timesteps, in comparison with larger ones. Conversely, Gurobi's success rate grows as the timesteps become larger. Although OpenWBO-lsu, NaPS, and Clasp do not command a significant portion of victories, they complement the behavior of the more dominant solvers within the portfolio.

Figure~\ref{matrix} summarizes the information on the best solver for each instance and for each timestep. In this figure, it is apparent that the best solver performance depends on the benchmark as well as the timestep. The clear implication is that there is no single best solver for all the instances and timesteps. Since instances belonging to the same family are plotted together, we can also observe that the behavior of the solvers in the portfolio seem to depend on the family of the instances. Most of the instances for which no feasible solution is found by any of the solvers across 500 timesteps belong to benchmarks ``mps-v2-20-10" and ``market-split'' from the 2006 version of the PB Competition, benchmarks ``opb-trendy" and ``opb-paranoid" from the PB Competition 2010 and PB Competition 2012. These benchmarks correspond to the competition's category called BIGINT, which means that the coefficients can be arbitrary precision (i.e. not bounded) integer numbers.

To train and test the ML models, the instances were partitioned into training and testing sets. This was done by partitioning the instances of the $118$ benchmarks into $70\%$ for training and $30\%$ for testing, resulting in $2054$ instances for training and $1074$ instances for testing. The partition was done by randomly picking the instances from each benchmark, maintaining the same ratio.

\subsection{Characterization and labeling}
\label{sub:characterization}

\subsubsection{Domain-specific features for PBO}
Based on previous work on SAT \citep{xu2008satzilla} and MaxSAT \citep{ansotegui2016maxsat}, we defined a set of features for our problem. For this selection, considering the anytime nature of our meta-solver, we focus on informative fast-to-compute features. Since our problem has its own characteristics, we also test some other features that are specific to non-linear PBO instances. Therefore, here we use the following $8$ sets of \textit{domain-specific features}:

\begin{description}
    \item[Number of constraints:] Number of constraints in the instance. An equivalent feature for SAT and MaxSAT was used by \citep{xu2008satzilla} and \citep{ansotegui2016maxsat}. 
    \item[Number of variables:] Number of Boolean variables present in the instance. This feature was used by \citep{xu2008satzilla} and \citep{ansotegui2016maxsat}. 
    \item[Linearity:] Identifies if the formula contains non-linear constraints. No similar feature was proposed before.
    \item[Distribution of the number of terms per constraint:] We partition the constraints into four classes according to the number of terms they contain: $1$, $2$, $3$, or $4$ or more terms. The four percentages of the number of constraints in each class out of the total are four features in this set. A similar set of features was used by \citep{xu2008satzilla}.
    \item[Term degree:] Percentage of unary, binary, ternary and quaternary-or-more terms. This is  the number of terms with $1$, $2$, $3$, or $4$ or more variables out of the total number of terms in the instance. No similar feature was proposed before.
    \item[Objective function size:] Percentage of terms that are present in the objective function, out of the total number of terms. No similar feature was proposed before.
    \item[Positive terms (Constraints):] Percentage of positive terms in the constraints. An equivalent feature was used by \citep{xu2008satzilla} and \citep{ansotegui2016maxsat}. 
    \item[Positive terms (Objective):] Percentage of positive terms in the objective function. Inspired by the above, we extend the feature for the objective function.
\end{description}

\subsubsection{Ground truth labeling for the models}
As mentioned in Section~\ref{sec:pbo-inst-solvers}, $7$ different solvers were used to create the meta-solver. We then use the solvers as labels to identify which solver is the best for a given instance-time pair. We include a ``no solution'' label to indicate the cases where no solver obtains a feasible solution at a given timestep. This can be useful, especially, for hard instances where solvers require a long time to compute the first feasible solution. Also, in practice, a ``no solution'' label may indicate to the user the need for allocating more computational resources for solving the instance. Nevertheless, this feature of our model was not used for the final evaluation of the meta-solver, since this kind of prediction is not usually considered in the $\hat{m}$ metric.

Despite the potential misalignment between accuracy and $\hat{m}$ metrics, we have chosen to train multi-label classification models that prioritize accuracy. This decision stems from the requirement of having simple and fast ML models for anytime scenarios. By employing a multi-label classification model, our approach offers the advantage of considering all solvers simultaneously and making a single call to the oracle to make a decision. This stands in contrast to more complex Algorithm Selection Systems that typically involve multiple ML oracles, such as multiple binary classification models for each pair of alternatives or regression models for individual solvers. The use of such complex systems would result in prohibitively long prediction times, which are not suitable for our anytime scenario.

\subsection{Machine Learning Models}
\label{sec:models}
\citep{loreggia2016deep} proposed a generic ML approach, using Convolutional Neural Networks (CNN), already described in Section~\ref{sec:related}, for designing an automatic algorithm selector able to work without the need for handcrafted domain-specific features. We test variants of this method against variants of Random Forest, Gradient Boosting and k-Nearest Neighbors based on the domain-specific characterization of Subsection~\ref{sub:characterization}.


\subsubsection{Models using domain-specific features}

In our study, we conducted experiments using three ML algorithms, as outlined in Section~\ref{sub:prelims-ml}. These algorithms were implemented using the Scikit-learn library \citep{scikit-learn}. We utilized various subsets of the 14 domain-specific features described in Section~\ref{sub:characterization}. Additionally, hyperparameter tuning was performed to determine the optimal architecture for each model, as well as weighting strategies to compensate for the natural bias induced by the dominating classes of the portfolio.

\begin{description}
    \item[RF\_basic:] The Random Forest classifier uses only two features: the number of constraints and the number of variables. The hyperparameters used were: \texttt{n\_estimators = 100}, \texttt{max\_features = "sqrt"}, \texttt{criterion = "gini"}.
    \item[RF\_nonlinear:] The Random Forest classifier uses all the $8$ sets of domain-specific features. The hyperparameters used were: \texttt{n\_estimators = 100}, \texttt{max\_features = "sqrt"}, \texttt{criterion="gini"}.
    \item[RF\_linear:] The Random Forest classifier uses features of  the linearized version of the PBO instance. Therefore, the features related to non-linearity and term degree are redundant and removed. The hyperparameters used were: \texttt{n\_estimators=100}, \texttt{max\_features = "sqrt"}, \texttt{criterion = "gini"}.
    \item[GB\_basic:] The Gradient Boosting classifier uses only two features: the number of constraints and the number of variables. The hyperparameters used were: \texttt{n\_estimators = 100}, \texttt{learning\_rate = 0.5}, \texttt{max\_depth = 3}, \texttt{max\_features="sqrt"}.
    \item[GB\_nonlinear:] The Gradient Boosting classifier uses all the $8$ sets of domain-specific features. The hyperparameters used were: \texttt{n\_estimators = 100}, \texttt{learning\_rate = 0.25}, \texttt{max\_depth = 3}, \texttt{max\_features = "sqrt"}.
    \item[GB\_linear:] The Gradient Boosting classifier uses features of  the linearized version of the PBO instance. Therefore, the features related to non-linearity and term degree are redundant and removed. The hyperparameters used were: \texttt{n\_estimators = 100}, \texttt{learning\_rate = 0.1}, \texttt{max\_depth = 3}, \texttt{max\_features = "sqrt"}.
    \item[KNN\_basic:] The $k-$Nearest Neighbors classifier uses only two features: the number of constraints and the number of variables. The hyperparameter used was: \texttt{n\_neighbors = 13}.
    \item[KNN\_nonlinear:] The $k-$Nearest Neighbors classifier uses all the $8$ sets of domain-specific features.  The hyperparameter used was: \texttt{n\_neighbors = 21}.
    \item[KNN\_linear:]  The $k-$Nearest Neighbors classifier uses features of  the linearized version of the PBO instance. Therefore, the features related to non-linearity and term degree are redundant and removed.  The hyperparameter used was: \texttt{n\_neighbors = 21}.
     
\end{description}

For all the variants, the set of features is augmented with the feature of timestep, which increments the number of the model's input features in one. We note that for the \texttt{linear} versions, we first need to linearize the input instance in order to compute the purely linear features, which is not the case for the \texttt{nonlinear} versions, for which we don't incur in such overhead for the computing of the non-linear features.

\subsubsection{CNN for Loreggia's Representation}


As a baseline method, we characterize the instances as images, following the proposal of \citep{loreggia2016deep} (described in Section ~\ref{sec:aasat}). These images are given as input to a Convolutional Neural Network (CNN), which outputs the best solver for every timestep. Hence, we adapt the method to handle anytime scenarios by learning a label for each of the possible $500$ timesteps. That way, when a prediction for a particular time is needed, we have to inspect the output of the network that corresponds to the closest (smaller or equal) timestep output.

For the implementation of the CNN, three different architectures were tested: \textbf{VGG16} \citep{simonyan2014very}, \textbf{AlexNet} \citep{krizhevsky2017imagenet} and \textbf{GoogLeNet} \citep{szegedy2015going}.

\subsection{Evaluation}
\label{sec:eval}

\begin{table}[h]
\centering
\begin{tabular}{lll}
 \hline \hline
 \textbf{ML Oracle} & \textbf{Accuracy}  & \textbf{Metric $\hat{m}$}  \\ \hline \hline
 Loreggia's w VGG & 0.4712 & 1.00 \\
 Loreggia's w AlexNet & 0.5775 & 0.7157 \\
 Loreggia's w GoogLeNet & 0.4931 & 1.3565 \\
 RF\_basic & 0.6580 & 0.7108 \\
 \textbf{RF\_nonlinear} & 0.7106 & \textbf{0.5250} \\
 \textbf{RF\_linear} & \textbf{0.7159} & 0.5729 \\
 GB\_basic & 0.6379 & 0.8252 \\
 GB\_nonlinear & 0.7046 & 0.6198 \\
 \textbf{GB\_linear} & \textbf{0.7184} & 0.6501 \\
 KNN\_basic & 0.6225 & 0.8481 \\
 KNN\_nonlinear & 0.6407 & 0.8589 \\
 KNN\_linear & 0.6621 & 0.7971 \\
 \hline
\end{tabular}
\caption{Accuracy and $\hat{m}$ values for different Machine Learning Models and subsets of characteristics.} 
\label{tab:eval}
\end{table}

Table~\ref{tab:eval} compares the accuracy and $\hat{m}$ values (calculated as described in \ref{sec:ms-perf}) of the different combinations of ML models and subsets of features as explained in the previous subsection. As can be seen, GB\_linear provides the best performance in accuracy and RF\_nonlinear the best performance in the $\hat{m}$ metric. 

The ML methods relying on domain-specific features, regardless of the subsets of features considered, outperform in accuracy the deep-learning networks based on the generic representation of \citep{loreggia2016deep}, which we take as a baseline. The Deep Learning Network that provides the best accuracy and $\hat{m}$ values is the one based on the AlexNet architecture.


Although not perfect, Table~\ref{tab:eval} demonstrates an inverse correlation relation between the accuracy and $\hat{m}$ metrics. This is with the noticeable exception of the best-performing Deep Learning Network, AlexNet, which, in comparison with GB\_basic and all KNN models, with a worse accuracy value achieves a better $\hat{m}$ score. 

It is important to note that this table does not account for the overhead associated with computing the features or the time required for the models to generate predictions. These factors can significantly impact the practical performance of using these models to build an anytime meta-solver. Therefore, we will further analyze and present results considering the four best-performing combinations of models and sets of features: RF\_nonlinear, RF\_linear, GB\_nonlinear, and GB\_linear. 

\begin{figure*}[h!]
    \subfloat[RF\_nonlinear]{%
        \includegraphics[width=.48\linewidth]{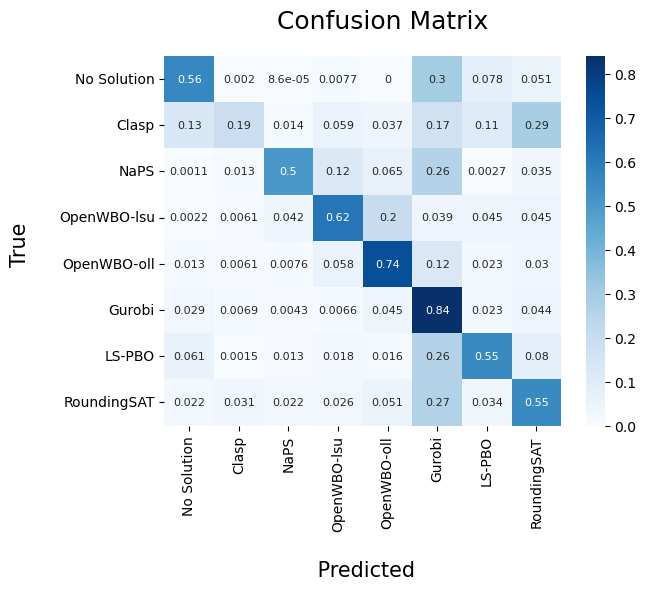}%
        \label{subfig:a}%
    }\hfill
    \subfloat[RF\_linear]{%
        \includegraphics[width=.48\linewidth]{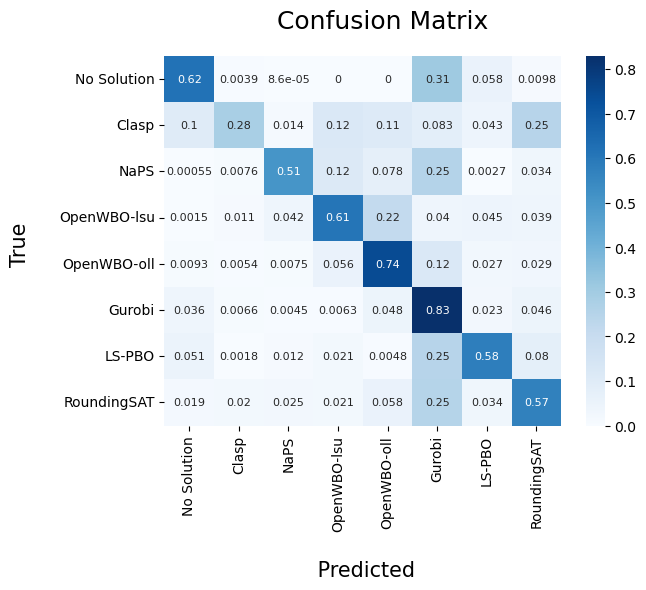}%
        \label{subfig:b}%
    }\\
    \subfloat[GB\_nonlinear]{%
        \includegraphics[width=.48\linewidth]{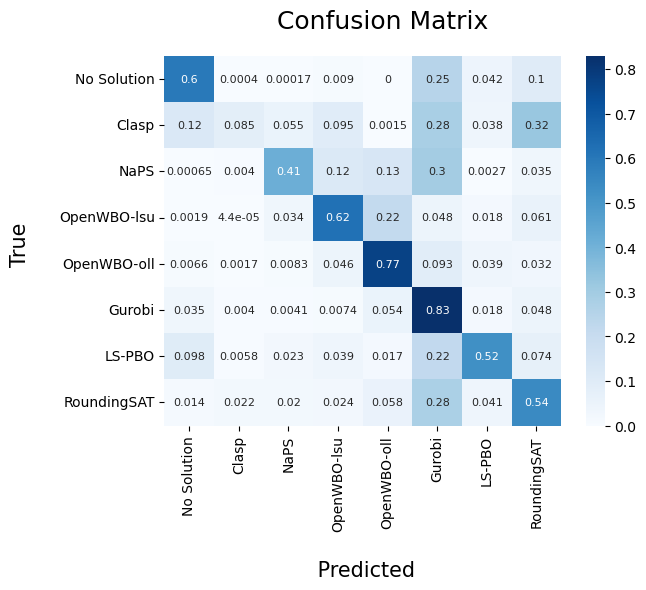}%
        \label{subfig:c}%
    }\hfill
    \subfloat[GB\_linear]{%
        \includegraphics[width=.48\linewidth]{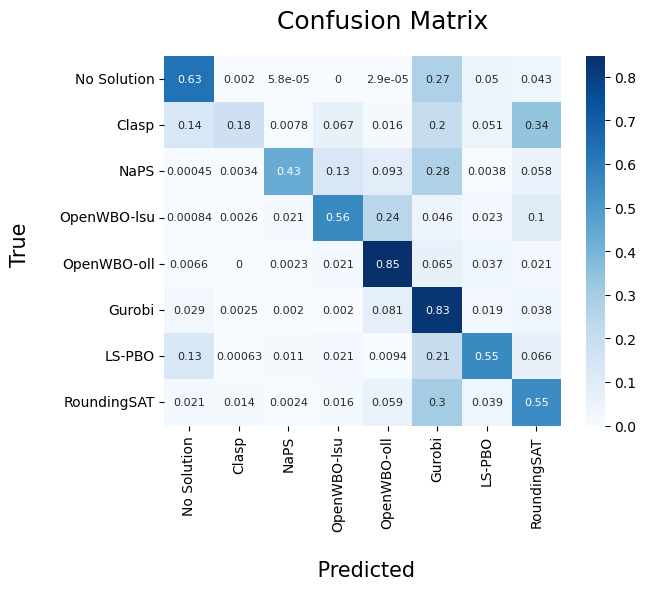}%
        \label{subfig:d}%
    }
    \caption{Confusion matrixes of PBO meta-solvers based on RF\_nonlinear, RF\_linear, GB\_nonlinear and GB\_linear.}
    \label{fig:conf_matrix}
\end{figure*}

Figure~\ref{fig:conf_matrix} depicts the Confusion Matrix for our best models. It is evident that all matrices demonstrate a similar pattern in the behavior of the models. Generally, it can be inferred that the classes were learnable, except for the Clasp class, which has a smaller representation in the dataset. It is natural for these models that the higher the class representation in the dataset, the higher the accuracy for that class. Similarly, a higher class representation increases the likelihood of the model over-predicting that class. To mitigate this issue, we implemented methods to address the bias introduced by the dominant classes. These methods involved assigning, during the training of the models, bigger weights to miss-classifications of less frequent classes, compared to the more dominant ones.

\begin{figure}[hhh!]
\centering
\begin{subfigure}{0.8\columnwidth}
\includegraphics[width=\columnwidth]{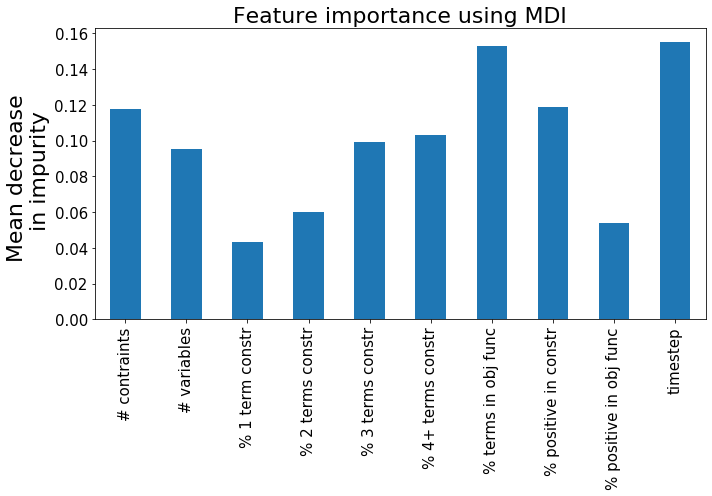}
    \caption{Feature importance for the RF\_linear model.}
    \label{subfig:feat_rf_nl}
\end{subfigure}
\par\bigskip
\begin{subfigure}{0.8\columnwidth}
\includegraphics[width=\columnwidth]{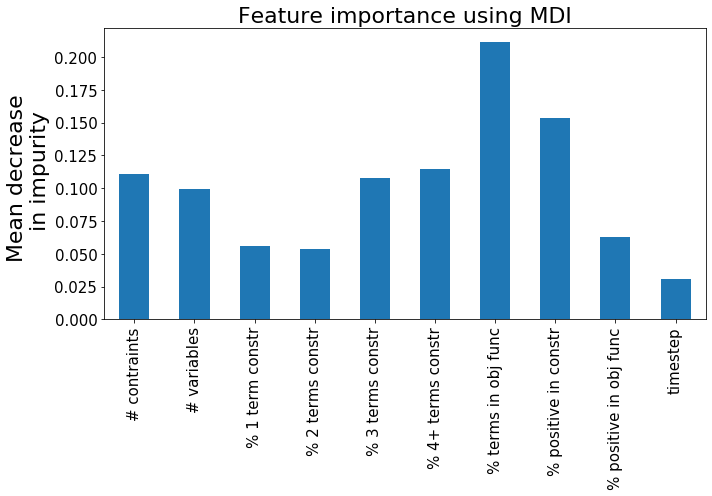}
   \caption{Feature importance for the GB\_linear model.}
    \label{subfig:feat_gb_nl}
\end{subfigure}
\caption{}
\label{fig:feat_import}
\end{figure}

The output of Random Forest and Gradient Boosting includes the MDI (Mean Decrease in Impurity) for each feature, which is a proxy for feature importance. The higher the value of the MDI, the more important the feature is. Figure~\ref{fig:feat_import} shows the MDI values for the $10$ features of the linearized instances for both models. It is evident that the importance of features varies depending on the model used. In particular, for the GB model, the timestep feature appears to be less significant compared to other features related to the composition of the PBO formula in making predictions. On the other hand, the RF model heavily relies on the timestep feature to make recommendations. Both models consider the \textit{percentage of terms that are present in the objective function}, first proposed here, as a very important feature.


\begin{figure}[h!]
\centering
\begin{subfigure}{0.9\columnwidth}
\includegraphics[width=\columnwidth]{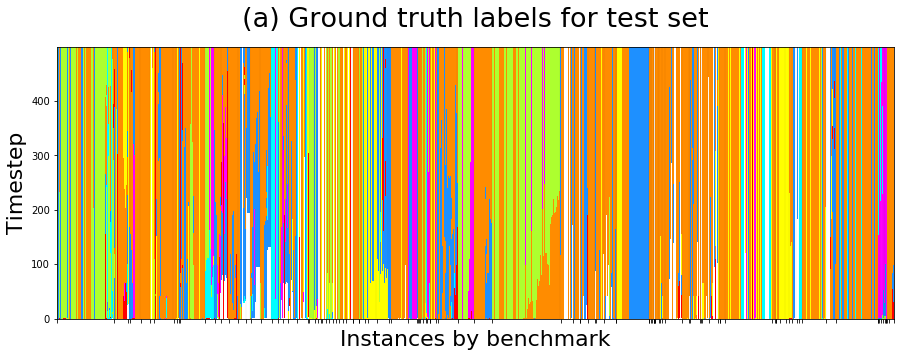}
\end{subfigure}
~\\
\begin{subfigure}{0.9\columnwidth}
\includegraphics[width=\columnwidth]{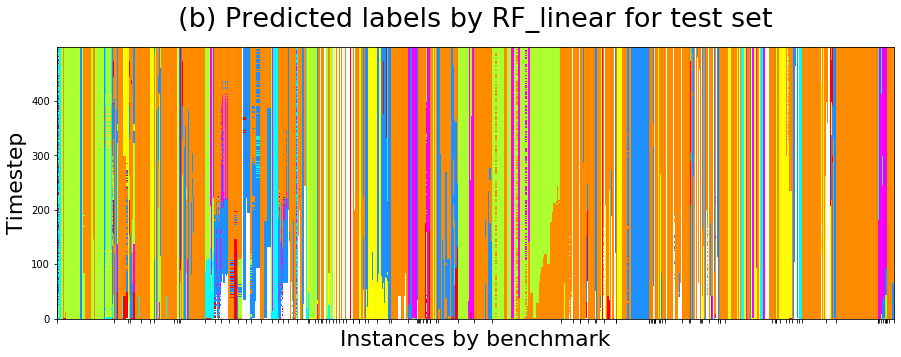}
\end{subfigure}
~\\
\begin{subfigure}{0.9\columnwidth}
\includegraphics[width=\columnwidth]{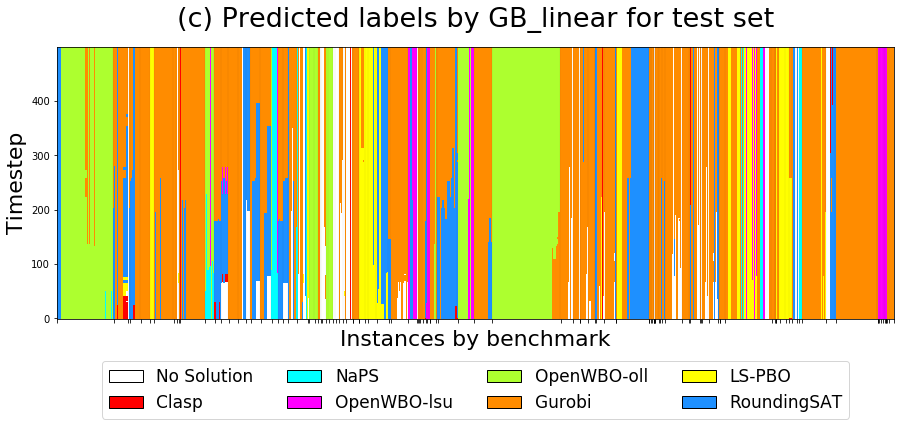}
\end{subfigure}
\caption{Comparison of ground truth labels with predicted labels of the RF\_linear and GB\_linear for the test set.}
\label{fig:contrast}
\end{figure}

Figure~\ref{fig:contrast} provides a visual representation of how our two most accurate models, RF\_linear and GB\_linear, behave. By examining this figure in conjunction with Figures~\ref{fig:feat_import},\ref{fig:conf_matrix}, and Table\ref{tab:eval}, we can draw some conclusions. 
Although GB\_linear achieves higher accuracy, this is primarily due to the bias introduced by the four most dominant classes, for which it exhibits superior performance compared to the RF model. Furthermore, it is evident that the GB model places less emphasis on the anytime behavior of the solvers and tends to select the same solver for a given instance, regardless of the timestep. In contrast, the RF model demonstrates a more varied selection of solvers based on the timestep. This observation aligns with the analysis of feature importance in the GB and RF models, providing support for this observation.

\section{Results}
\label{sec:results}
\subsection{Meta-solver's performance}
\label{sec:ms-perf}
In this section, we present the results of the performance of our meta-solver based on the four best models explained in the previous Section: RF\_nonlinear, RF\_linear, GB\_nonlinear, and GB\_linear ML oracles. As explained in Subsection~\ref{sub:aas}, the best way to measure the performance of a meta-solver is through the $\hat{m}_{ms}$ metric. 
For our particular case, $\hat{m}_{ms}$ was calculated considering Gurobi as the SBS, for all timesteps.
For computing the cumulative score $m_s$, for each solver $s$, as defined in Equation~\ref{eq:m-s}, the normalization function $n(o_s(i,t),i,t)$ of $o_s(i,t)$ (the objective value of $s$ on instance $i$ at timestep $t$),  is defined so that its co-domain is in the range $[0,1] \cup \{2\}$. For this, we compute $o_{min}(i)$, the minimum feasible value (in many cases the optimal value) of the objective function for the instance $i$ and $o_{max}(i)$, the maximum feasible value of the objective function for the instance $i$, both considering all the feasible solutions found by all the solvers. The by-default normalization of a given value $o_s(i,t)$ is computed as follows:
 \begin{equation}
  \label{eq:n1}
    n'(o_s(i,t),i,t) = \dfrac{o_{s}(i,t) - o_{min}(i)}{o_{max}(i) - o_{min}(i)}
 \end{equation}

This by-default normalization is not always well defined and some special cases have to be considered. Considering such cases, we formally define $n(o_s(i,t),i,t)$ as:
 \begin{equation}
 \label{eq:n}
        \begin{cases}
            0 & \text{if }o_s(i,t) = o_{min}(i) = o_{max}(i) \\
            2 & \text{if }o_s(i,t)\text{ is undefined}\\
            &\text{but }o_{max}(i)\text{ is defined} \\
            n'(o_s(i,t),i,t) & \text{  otherwise}
        \end{cases}
\end{equation}

As $\hat{m}_{ms}$ compares the meta-solver with SBS, and such solver is not able to use the ``no solution" label in its favor, for our meta-solver's evaluation, we decide to only consider instance-time pairs for which $o_{max}(i,t)$ is defined (i.e. we don't consider the instance-timesteps pairs that correspond to white points on Fig~\ref{subfig:ground-truth}).

One issue to consider concerning the computational time limit is whether to include the feature computation and prediction times needed by the ML Oracle in addition to running the solver. The prediction requires input preparation for the instance (computing the features for the model, constructing the image for CNN, and linearizing the instances for the models with only linear features) and running the ML model. If we consider the prediction time, there is less time to run the solver and, consequently, the value of our performance metric $\hat{m}_{ms}$ goes up. We report on the performance of RF\_nonlinear, RF\_linear, GB\_nonlinear, and GB\_linear for both cases, when prediction time ``overhead" is included or not, for each timestep, in Table~\ref{tab:m-hats-overhead}.

\begin{table}
\centering
\begin{tabular}{lll}
 Model & $\hat{m}_{ms}$ (no) &  $\hat{m}_{ms}$ (o)
 \\ \hline \hline 
 RF\_nonlinear & \textbf{0.5250} & \textbf{0.5318}  \\ 
 RF\_linear & 0.5729 & 0.6042 \\
 GB\_nonlinear & 0.6198 & 0.6270 \\ 
 GB\_linear & 0.6501 & 0.6712 \\
\end{tabular}
\caption{$\hat{m}_{ms}$ calculated with no overhead time (no) and $\hat{m}_{ms}$ calculated considering the overhead time (o) for the PBO meta-solvers based on RF\_nonlinear, RF\_linear, GB\_nonlinear and GB\_linear. Lower $\hat{m}_{ms}$ values are better. } 
\label{tab:m-hats-overhead}
\end{table}

It is evident that while the RF\_linear model exhibits the better accuracy value, it is the RF\_nonlinear one that achieves the best $\hat{m}$ value. The difference in the $\hat{m}$ scores between these two models grows even bigger when considering the overhead. This is primarily due to the time impact of linearizing the instances before computing the features in the RF\_linear case. This also happens with the GB models, although the $\hat{m}$ values are less competitive than the ones achieved by the RF models. 

\begin{figure}[h]
    \centering
    \includegraphics[width=1\columnwidth]{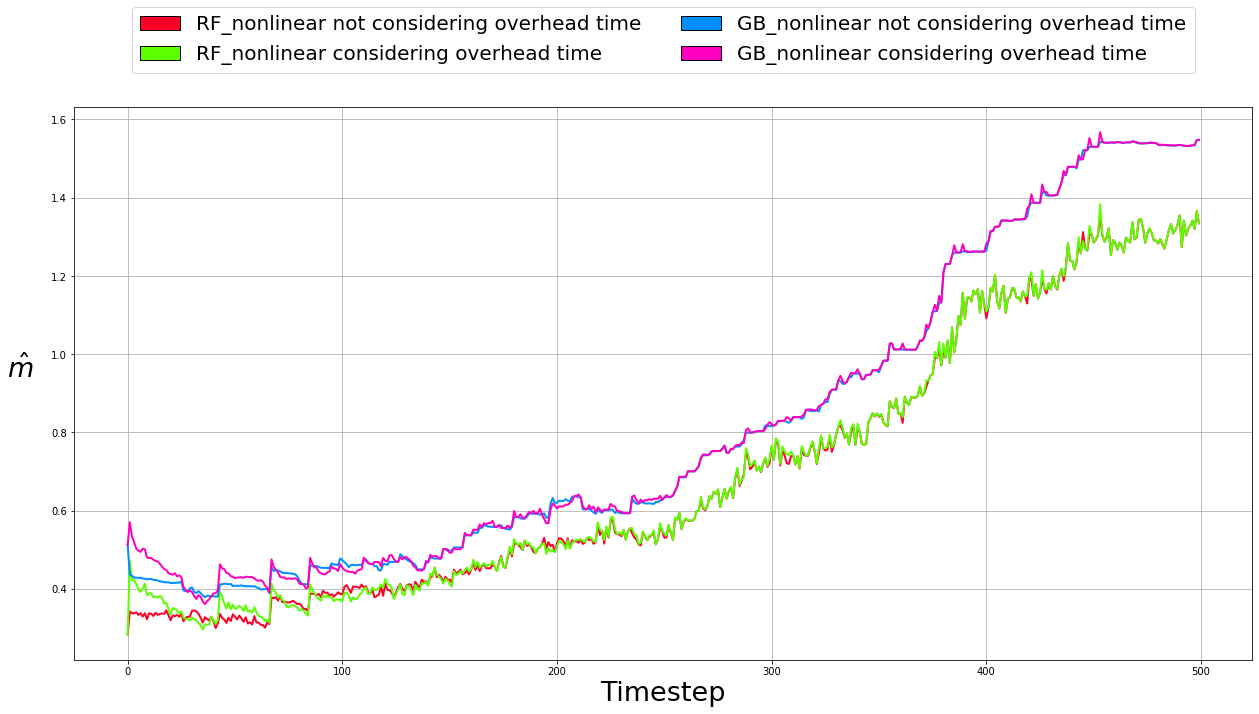}
    \caption{Values of $\hat{m}$ across timesteps for GB\_nonlinear and RF\_nonlinear. Lower $\hat{m}$ values are better.}
    \label{fig:m-hat-timesteps}
\end{figure}

Figure~\ref{fig:m-hat-timesteps} provides insight into how the $\hat{m}$ value changes for each timestep for the best-performing RF and GB models, taking into account the overhead. As anticipated, we observe that the overhead has a negative impact on the $\hat{m}$ value during the initial timesteps. RF consistently demonstrates better $\hat{m}$ values than GB, suggesting that RF learns more effectively from the anytime data.



\subsection{Comparing the meta-solver with the Single Best Solver}
\begin{figure}[h]
    \centering
    \includegraphics[width=0.8\columnwidth]{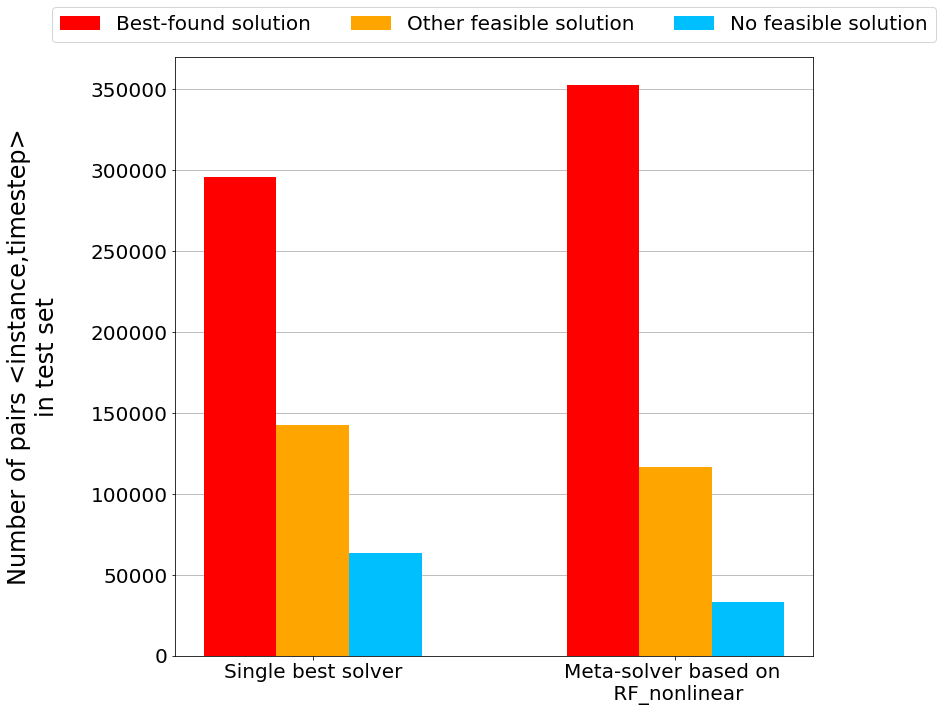}
    \caption{Number of feasible instance-timestep pairs of the test set where the SBS (Gurobi), and the meta-solvers based on RF\_nonlinear and GB\_nonlinear find the optimal solution, feasible non-optimal solution or no feasible solution.}
    \label{sbs-solutions}
\end{figure}

Recall that Gurobi is the single best solver (SBS).
Here we elaborate on where the gain from the meta-solver (MS) comes from. To do this, we consider all \textit{test} instance-timestep pairs for which a feasible solution is known, $502258$ in total. In Figure~\ref{sbs-solutions}, we compare, over this set of instances, the number of instances for which each of the solvers (SBS and the MS based on the RF\_nonlinear model) report the best-found incumbent solution (red), a feasible worse-than-the-best-found incumbent solution (orange) or for which no incumbent solution has been computed yet (light blue). It is apparent from this figure that the meta-solver MS provides a significant improvement over the use of the SBS by, for many instance-timestep pairs, selecting alternative solvers that are either able to find better solutions than the SBS or that are able to compute an incumbent solution when the SBS is not. This justifies the use of our meta-solver for the PBO problem in practice. 

Overall, from $502,258$ test instance-timestep pairs, Gurobi is able to find $296,103$ best-found solutions, $142,824$ non-best-found incumbent solutions, and is unable to find feasible solutions for $63,331$ instance-timestep pairs. The meta-solver finds $352,462$ best-found solutions, $116,717$ non-best-found solutions and is unable to find feasible solutions for $33,079$ instance-timestep pairs. As we can see, the meta-solver improves Gurobi's performance by achieving the best-found solution in around $19\%$ more instance-timestep pairs and diminishing in up to $47.7\%$ the number of instance-timestep pairs for which a feasible solution is not yet found.

\section{Conclusions and Future Work}
\label{sec:conclusions}
We propose here an Anytime meta-solver for the Pseudo-Boolean Optimization problem.  Our meta-solver is able to predict and execute a solver that, among 7 different solvers, performs best for a given problem instance and a specified time limit. Our results show that our meta-solver (based on any of the two best models) significantly outperforms all individual solvers, while it also identifies when feasibility cannot be achieved for a given instance. 


A logical next step is to propose ways of adapting Anytime Algorithm Selection Problems to the scenarios of the Algorithm Selection Library \citep{bischl_aslib_2016}, which, currently, are not \textit{anytime}. We will use this as an efficient way of sharing our data.

For future work, we plan to explore the application of Graph Neural Networks as a potential ML oracle. This type of Neural Network has recently been shown to perform well on data that can be represented as a graph, which is the case for PBO.  We also plan to explore a two-layer meta-solver approach, where the first layer selects a solver from a portfolio while the second layer chooses the most suitable set of parameters for the chosen solver. 



\section*{Acknowledgments}
First, second and last authors are supported in part by AI institute NSF award 2112533. 


\end{document}